\documentclass[10pt,twocolumn,final]{IEEEtran}
\usepackage[dvips]{graphicx}
\usepackage[para]{threeparttable}
\usepackage{amsmath}
\usepackage[noadjust]{cite}
\ifCLASSINFOpdf
\else
\fi
\begin{document}
%
\title{Scale Selection of Adaptive Kernel Regression by Joint Saliency Map for Nonrigid Image Registration}
%
%
%
%

\author{Zhuangming Shen, Jiuai Sun, Hui Zhang, and Binjie Qin*,~\IEEEmembership{Member,~IEEE}
\thanks{Manuscript received February 20, 2013; revised April 26, 2013; This work was supported in part by NSFC (61271320, 60872102 and 60402021), NBRPC (2010CB834300), China Scholarship Council and the small animal imaging project (06-545).}
\thanks{Zhuangming Shen is with the School of Biomedical Engineering and Med-X Research Institute, Shanghai Jiao Tong University, Shanghai, 200240, China (e-mail: shenzhuangming@gmail.com).}
\thanks{Jiuai Sun is with the Machine Vision Laboratory, University of the West of England, Bristol, BS16 1QY, U.K. (e-mail: jiuai2.sun@uwe.ac.uk).}
\thanks{Hui Zhang is with Centre for Medical Image Computing, University College London, WC1E 6BT London, U.K. (e-mail: g.zhang@cs.ucl.ac.uk).}
\thanks{*Binjie Qin is with the School of Biomedical Engineering and Med-X Research Institute, Shanghai Jiao Tong University, Shanghai, 200240, China. During 2012 to 2013, he is also with Centre for Medical Image Computing \& Department of Medical Physics and Bioengineering, University College London, WC1E 6BT London, U.K. (e-mail: bjqin@sjtu.edu.cn).}}

%
%

\markboth{Journal of \LaTeX\ Class Files,~Vol.~6, No.~1, January~2013}%
{Shell \MakeLowercase{\textit{et al.}}: Bare Advanced Demo of IEEEtran.cls for Journals}
%



\IEEEcompsoctitleabstractindextext{%
\begin{abstract}
Joint saliency map (JSM) \cite{1} was developed to assign high joint saliency values to the corresponding saliency structures (called Joint Saliency Structures, JSSs) but zero or low joint saliency values to the outliers (or mismatches) that are introduced by missing correspondence or local large deformations between the reference and moving images to be registered. JSM guides the local structure matching in nonrigid registration by emphasizing these JSSs' sparse deformation vectors in adaptive kernel regression of hierarchical sparse deformation vectors for iterative dense deformation reconstruction. By designing an effective superpixel-based local structure scale estimator to compute the reference structure's structure scale, we further propose to determine the scale (the width) of kernels in the adaptive kernel regression through combining the structure scales to JSM-based scales of mismatch between the local saliency structures. Therefore, we can adaptively select the sample size of sparse deformation vectors to reconstruct the dense deformation vectors for accurately matching the every local structures in the two images. The experimental results demonstrate better accuracy of our method in aligning two images with missing correspondence and local large deformation than the state-of-the-art methods.

\end{abstract}

}

\maketitle

\IEEEdisplaynotcompsoctitleabstractindextext

%
\IEEEpeerreviewmaketitle

\section{Introduction}
%
%

%
%
%
%
\IEEEPARstart
{N}onrigid image registration \cite{2} is a procedure to minimize the difference between one (reference) image and another (moving) image by spatially aligning every corresponding local structures. In early detection of pathology, correctly matching corresponding local small structures is especially important to identify differences in morphology which are distinctive between pathological and heathy states. However, owing to the the outliers introduced by missing correspondence, such as the tumor appearing in preoperative image but not the intraoperative image, and/or local large deformations in the two images to be registered, robustly determining the accurate one-to-one correspondence between local structures is still a challenging task. Especially, these missing correspondences of local structures are always accompanied by the local large deformations.

These outliers exhibit large structural discrepancies in the local varying spatial context, where the mismatches between the local structure pairs could be so complex that the surrounding local structures could distort in very different ways. Due to the outlier presenting these various local differences, the local model \cite{3}\cite{4}\cite{5} based methodology that can deal with the locally varying difference are considered as an ideal methodology to account for these outliers. To tackle these outlier problems, the image registration methods that are classified into intensity- and feature-based registration methodologies have seen various efforts in recent years. Using local model of sparse image representation to select some corresponding features of two images, feature-based registration can be considered as local model based registration to directly matching the local structures by finding a geometric transformation from these sparse feature correspondences. However, the computation of registration is still sensitive to the false correspondences in the outliers. Recently, a Bayesian regression model \cite{6} successfully infer the continuous and locally smooth transformation for registering challenging 2D point sets and was favorably compared with state-of-the-art methods \cite{7}\cite{8}\cite{9} both in cases of noise and outliers. This work justifies that local model (such as regression model) are favorable to model registration transformation compared with other interpolation based techniques.

We regard most intensity-based registration as global model based registration that is often formulated as a global energy minimization problem with the energy being composed of an regularization term and a similarity term. Due to the global model driven whole-intensity similarity being unable to represent the similarity of the local structures, the outlier problems were only partially solved by using a locally varying weight between regularization and similarity \cite{10}, creating artificial correspondence \cite{11}, or removing the outliers with cost-function masking \cite{12}. These approaches either are largely dependent on the outlier segmentation or not automatically tackle the missing correspondence and local large deformations simultaneously.

Our previous work proposed a joint saliency map (JSM) \cite{1} to highlight the corresponding saliency structures (called joint saliency structures, JSSs) in the two images, and emphatically group those JSSs in the weighted joint histogram computation for the automatic rigid intensity-based registration of challenging image pairs with outliers. After getting the sparse deformation vectors of moving image in a hierarchical block matching, we further use the JSM to emphasize these JSSs' sparse deformation vectors in the JSS adaptive kernel regression for automatically reconstructing dense deformation vectors in intensity-based nonrigid image registration \cite{13}. The local structures' registration (deformation) accuracy in local estimates is mainly dependent on the shape/size of the neighborhood deformation vectors and/or the estimation weights used for local estimation. Our JSS adaptive kernel regression adapts the kernel function's shape and orientation to the reference image's local saliency structure, more displacement vector samples belonging to the same local structure are grouped together so that the regression of local deformation can accord with the local saliency structures in the reference image. In addition, JSM highlights the weights of JSSs at the moving windows/kernel in reconstructing local dense deformation vectors while suppressing outlier effects in the regression.

However, our method still could not accurately describe the deformation in some local small structures because the windows size for the kernel regression is fixed. The scale (width) of moving window/kernel determines the sample size of sparse displacement vectors participating in the kernel regression and therefore controls the amount of deformation smoothing introduced by the local approximation. A small scale means a small window and corresponds to noisy estimates, less biased, and with high variance. Comparatively, a large scale corresponds to a large window and therefore to smooth deformation estimates, with low variance and typically increased estimation bias. Thus, the local scale of kernel regression controls the trade-off between the registration accuracy and the smoothness of the local deformation field. The optimal choice of kernel scale depends on the mismatch degree (registration inaccuracy) and structural scale of underlying local structures to be matched. For large structures and large mismatches, we would like the kernel scale to be large to reduce the registration (or deformation) variance. For small structures and small mismatches, a small kernel scale is desirable in order to reduce the registration bias error. Therefore, the local scale of kernel regression should be adaptively proportional to the structure scale and the mismatch degree (scale) of underlying saliency structures to be registered.

With the above-mentioned observations in mind, we propose a new method which has three contributions. 1) We propose mismatch scale into the nonrigid image registration by using JSM, whereby we could judge the registration inaccuracy in the local structure matching. 2) We design a simple but effective superpixel based local structure scale estimator, which first segments the reference image into multi-resolution superpixel \cite{30}\cite{18} structural regions and then calculate the structure scales of Gaussian smoothed superpixel regions in terms of variance in a scale-space framework through the minimal description length criterion (MDL) \cite{19}\cite{35}. 3) We introduce a local kernel scale selection scheme by conflating the mismatch scale with the superpixel based structure scale, and apply it to our previous nonrigid image registration using JSS adaptive kernel regression. By integrating this local scale selection scheme into multi-resolution adaptive kernel regression, the nonrigid registration can iteratively guide the deformation of each local structure towards the well-aligned position and orientation. Therefore, we can achieve more accurate local structure matching in small structures and maintain a smooth deformation field around local structures. The proposed method is elaborated in Section 2 followed by experimental results in Section 3. The whole paper is concluded in Section 4.

\section{Background and Related Works}
\subsection{Joint Saliency Structure Adaptive Kernel Regression}
Inspired by the success of local approximation by kernel regression (or nonparametric regression) for signal reconstruction, we consider the nonrigid image registration as a local adaptive kernel regression by iteratively reconstructing dense deformation vectors from the sparse deformation vectors obtained through hierarchical block-matching. After Suarez \emph{et al}. \cite{14} used the normalized convolution \cite{15} to estimate dense deformation field from sparse deformation field, two recent works \cite{16}\cite{17} also utilized kernel regression to estimate registration transformation. However, these methods did not exploit the local adaptivity of kernel regression for the nonrigid image registration with outliers.

Suppose we have sparse and irregularly distributed deformation vectors $\{\mathbf{y}_i,\mathbf{x}_i\}_{i=1}^{P}$ given in the form
\begin{equation}
\label{eq1}
\mathbf{y}_i = \mathbf{z}(\mathbf{x}_i)+\mathbf{\varepsilon}_i,\quad \mathbf{x}_i\in\Omega,~~i=1,\cdots,P
\end{equation}
where the $\mathbf{y}_i$ is a sparse displacement vector (response variable) at position (explanatory variable) $\mathbf{x}_i$, $\mathbf{z}\left({\cdot}\right)$ describes the desired dense deformation field in the moving windows (kernel) $\Omega$ with independent and identically distributed zero mean noise ${{\varepsilon }_{i}}=\varepsilon \left( {{\mathbf{x}}_{i}} \right)$. In statistics, the function $\mathbf{z}\left({\cdot}\right)$ is treated as a regression of $\mathbf{y}$ on $\mathbf{x}$, $\mathbf{z}\left( \mathbf{x} \right)=E\left\{ \left. \mathbf{y} \right|\mathbf{x} \right\}$. In this way, the reconstruction of nonrigid deformation field is from the field of the regression techniques.

Importantly, our JSS adaptive kernel regression has first proposed two local adaptivity in selecting local kernel's shape and the JSS-based weights within moving kernel for local estimation. The workflow of JSS adaptive kernel regression combined with kernel scale selection for nonrigid registration is illustrated in Fig. 1, where different levels have their own resolution but the same procedure. At each level, the resulted deformation is composed of initial deformation and current deformation. The proposed method consists of an iterative scheme, which at each iteration alternates between the block matching and JSS adaptive kernel regression with local scale calculation. Firstly, we learn the underlying characteristics of sub-blocks' similarities to get roughly registered moving image's sparse displacement vectors. Then we compute the JSM of two images to highlight the locally JSSs and estimate the local scale of mismatch between the underlying saliency structures for subsequent kernel regression. Furthermore, we estimate every reference structure's orientation to design anisotropic kernel, and conflate the reference structure's structure scale with the mismatch scale for the selection of kernel scale. Finally, with a moving window/kernel in kernel regression, the output dense deformation vectors are estimated based on an emphatical weighting of the JSSs' sparse deformation vectors within the moving kernel with local adaptive scale (window size). Compared with our previous works \cite{13}, the proposed local adaptive scale selection for kernel regression is displayed at the module within the red dashed line in Fig. 1.
\begin{figure}[!t]
\centering
\centerline{\includegraphics[width=3.5in]{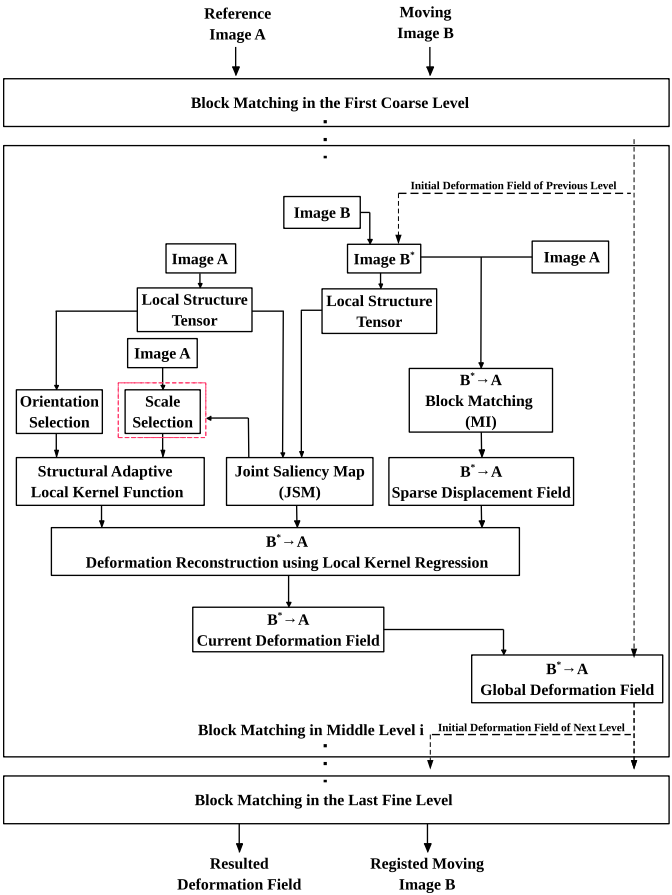}}
\caption{\small{\textbf{Flowchart of our algorithm in a coarse-to-fine framework}}}
\label{fig1.eps}
\end{figure}

\subsection{Local Scale Selection}
The idea of scale (window size, or bandwidth) selection is not new for kernel regression (or nonparametric regression). The scale parameter controls the trade-off between bias and variance in the local estimation of kernel regression. The are two types of approaches which have been reported over the last decades for scale selection in kernel regression \cite{4}\cite{22}. One is the \emph{plug-in} methods which calculate the ideal scale by estimating the bias and the variance under the mean squared error (MSE) between the real signal and its approximation \cite{18}\cite{19}. Alternatively, the \emph{goodness-of-fit} methods \cite{20}\cite{21}\cite{22} are widely used as data-driven methods without the bias estimates. These methods choose scale based on the accuracy criteria, with the main goal to achieve an optimal accuracy balancing the bias and the variance of estimation.

Because the kernel scale for JSS adaptive kernel regression are used to minimize the differences between the local saliency structures in the two images by selecting appropriate size of the neighborhood sparse deformation vectors, it is reasonable that the kernel scale should be consistent with the degree of mismatch (or local deformation) between the underlying saliency structure pairs and the scale of the saliency structures to be registered. Therefore, the kernel scale is adaptively estimated by combining the local mismatch scale with the structure scale of underlying saliency structures. Due to the \emph{goodness-of-fit} methods having capability for scale selection for image intensities and their various derivatives, we use this accuracy-based scale selection to estimate the kernel scale for JSS adaptive kernel regression.

To derive some accuracy criteria for selecting the local mismatch scale, we should first automatically determine the local registration inaccuracy (or registration uncertainty) during the nonrigid image registration procedure. Recently, locally evaluating the intensity-based nonrigid registration inaccuracy is a special subject that obtained increase research concerning (see \cite{23}-\cite{28} and references therein). Assuming the transformation parameters follow some prior statistical distributions, most of these methods search the entire image to infer the distribution of probable registration transformation. Due to the outliers presenting locally varying differences in the spatial contexts, this global model based prediction of registration transformation would have its limitations in accurately matching the local structures. Being different from these methods, an iterative JSM-based local strategy is presented, which is able to represent the degree of matching (or mismatch) of local saliency structures by computing the local similarity measures between the saliency-based \cite{29} appearance distributions on pixel pairs in the two images.

In fact, the JSM \cite{1}\cite{13} assigns high joint saliency values to the JSSs but zero or low joint saliency value to the mismatch structures (or regions) in the two images, whereby JSM is an ideal tool to indicate the local registration inaccuracy. Specifically, the moving saliency structures that require large deformations to be matched with reference structures are in the noncorresponding regions rather than the JSS regions, therefore their joint saliency values are low in JSM and the local kernel should be expanded to gather more sparse displacement vector samples for the right deformation. Relatively, the moving saliency structures that require zero or small deformation are in the JSS regions, where the joint saliency values are high in JSM and a small kernel can already determine the well-aligned deformation accuracy. Therefore, we adopt the JSM to design the mismatch scale for selecting kernel scale for JSS adaptive kernel regression.

We further investigate the selection of structure scale in the multi-resolution block matching based registration framework, which is preferred for modeling local large deformation of local saliency structures. Being a compact representation of an original image in low resolutions, large saliency structural regions initialize the sparse deformation vectors while the small local structures gradually refine the sparse displacement vectors with the increasing resolution of images. At the same resolution level, matching the local large structures could use large kernel scale to reduce the deformation variance (or increase deformation smoothness) compared with matching local small structures by using small kernel scale to reduce registration (or deformation) bias error. This means that, the kernel scale should be large for the large structural regions to adequately smooth deformation field and be small for the small structural regions to accurately match image details.

With this general scheme in mind, we use multi-resolution superpixel \cite{30}\cite{18} representation for their preserving structure boundaries to hierarchically segment the reference image into different saliency structural regions. The different scales for the different saliency structures in multi-resolution could be computed, by the local amount of Gaussian smoothing within the superpixel-represented saliency regions, in terms of variance in a scale-space framework.

\section{Methods}
\subsection{Structure Scale Selection}
Firstly, we use an accurate superpixel representation called Simple Linear Iterative Clustering (SLIC) \cite{30} to segment the reference image $I_R(\mathbf{x})$ into several small structural regions (superpixels) that adhere to the saliency structure boundaries in the reference image. Therefore, \emph{these structural regions are the representation of underlying local saliency structures}. Denote the image region as $\Phi$, and local structures as $S_i, i=1,\cdots,n$. Obviously, $\Phi=\bigcup_{i=1}^{n}S_i$.

Simultaneously, a discrete scale space is constructed by diffusing $I_R(\mathbf{x})$ with anisotropic diffusion equation \cite{31}
\begin{equation}
\begin{split}
\label{eq2}
\frac{dI_{R_{\sigma_k}}(\mathbf{x})}{dt} = C(\mathbf{x})\triangle I_R(\mathbf{x})+\nabla C(\mathbf{x})\nabla I_R(\mathbf{x}) \\
\end{split}
\end{equation}
where $C(\mathbf{x})=\exp(-(\frac{\parallel\nabla I_R(\mathbf{x})\parallel}{K})^2)$ is the diffusion coefficient, and the subscript $\sigma_k$ means a certain scale from a scale set $\Sigma=\{\sigma_1,\cdots,\sigma_m\}$. In this work, we assume the largest scale in the scale set is 15 pixels and the smallest one is 1 pixel. Because an image can be decomposed into a smoothed component and a residual component through anisotropic diffusion filter, the intensity on local superpixel $S_i$ can be represented by the smoothed component $I_{R_{\sigma_k}}(S_i)$ and a residual component $\varepsilon_{\sigma_k}(\mathbf{x}) = I_R(\mathbf{x}) - I_{R_{\sigma_k}}(\mathbf{x}), \mathbf{x}\in S_i$. The residual component can be modeled as a random field with zero-mean Gaussian density. The objective of local structure scale selection is to assign a scale $\sigma_k, k\in\{1,\cdots, m\}$ from $\Sigma$ for each local structure $S_i$ such that the posterior probability $P(\sigma_k|S_i)=\frac{P(\sigma_k)P(S_i|\sigma_k)}{P(S_i)}\propto P(S_i|\sigma_k)=\prod p(\mathbf{x}|\sigma_k), \mathbf{x}\in S_i$, achieves maximum, where $p(\mathbf{x}|\sigma_k)=P(I_R(\mathbf{x})|\sigma_k)$ is the likelihood of the observed image at each pixel $\mathbf{x}$ at scale $\sigma_k$, the $P(S_i|\sigma_k)$ is the likelihood of the observed structural region $S_i$ at scale $\sigma_k$.

To estimate a likelihood of the observed image at each pixel, we use the well known MDL criterion \cite{35} to relate the probability of an item with the length of the ideal code used to describe it, namely:
\begin{equation}
\label{eq3}
P(I_R|\sigma_k)=2^{-L(I_R|\sigma_k)}
\end{equation}
where $L(I_R|\sigma_k)$ denotes the description length of $I_R$ based on its decomposition at scale $\sigma_k$. This description length can be expressed as $L(I_R|\sigma_k)=L(I_{R_{\sigma_k}})+L(\varepsilon_{\sigma_k})$. Furthermore, the description length of the smoothed component $L(I_{R_{\sigma_k}})$ is assumed \cite{19} to be inversely proportional to the $\sigma_k^2$ while the description length of the residual component $L(\varepsilon_{\sigma_k})$ being proportional to the $\varepsilon^2_{\sigma_k}$, the $p(\mathbf{x}|\sigma_k)=P(I_R(\mathbf{x})|\sigma_k)$ can be estimated by
\begin{equation}
\begin{split}
\label{eq4}
\hat{p}(\mathbf{x}|\sigma_k) = A\mathbf{e}^{[-B(\frac{C}{\sigma^2_k}+\varepsilon^2_{\sigma_k}(\mathbf{x}))]}, \mathbf{x}\in S_i \\
\end{split}
\end{equation}
where $A$ is the normalizing constant, $B>0$ and $C>0$ are the empirical parameters adjusting the impact of the smoothed component and the residual component, which are set to 1 in this work.

By considering the scale coherence between neighboring local structures, the Markov Random Field (MRF) model is also implemented in this structure scale selection \cite{18}\cite{19}. As a result, the final structure scale selection is defined as
\begin{equation}
\begin{split}
\label{eq5}
& \sigma_s = \arg\max_{\sigma_k}P(\sigma_k|S_i)+\\
&~~~~~~~\lambda\sum_{\langle i,j\rangle}\delta(\sigma_k,\sigma_l)\exp(-(\mu(S_i)-\mu(S_j))^2) \\
& where  \\
& \delta(\sigma_k,\sigma_l) =
\begin{cases}
1, &if~\sigma_k = \sigma_l
\cr 0, &otherwise
\end{cases}
\end{split}
\end{equation}
$\mu(S_i)$ and $\mu(S_j)$ are the mean intensity on $S_i$ and $S_j$, respectively. $\sigma_k$ and $\sigma_l$ are the scales on $S_i$ and $S_j$ from the scale set $\Sigma$. In equation (4), the first term is the posterior probability on $S_i$, the second term is a smoothness function of the local structure $S_i$ and its neighboring local structure $S_j$. The second term prefers same scale labeling for neighboring pairs of appearance-similar superpixel regions, and penalizes same scale labeling between neighboring pairs of appearance-different superpixel regions. The impact of MRF is controlled by the parameter $\lambda$, which needs to be set to a small value (0.05) in order to avoid the over-smoothness that unintentionally increases the structure scales of the local small structures.

\subsection{Mismatch Scale Calculation using JSM}
According to our previous work \cite{13}, we define a center-surround saliency operator based on the contrast among neighboring local structure tensors (LST). This contrast emphasizes the dissimilarity or discrepancy between neighboring local structure tensors. For a given point $\mathbf{x}_0$ and its neighborhood $\Theta$, the saliency value $S(\mathbf{x_0})$ at $\mathbf{x}_0$ in a salient map can be computed through
\begin{equation}
\label{eq6}
S(\mathbf{x_0})=\mathrm{avg}\sum_{\mathbf{x}\in\Theta}\|\text{LST}(\mathbf{x})-\text{LST}(\mathbf{x}_0)\|_D
\end{equation}
where $\|\cdot\|_D$ defines a distance metric describing the dissimilarity between two LSTs. The operator avg computes the average of the dissimilarities within the neighborhood $\Theta$ of $\mathbf{x}_0$. The distance metric \cite{32} between two tensors $\mathbf{T}_1$ and $\mathbf{T}_2$ can be expressed as
\begin{equation}
\label{eq7}
\|\mathbf{T}_1-\mathbf{T}_2\|_D=\sqrt{\frac{8\pi}{15}(\|\mathbf{T}_1-\mathbf{T}_2\|_C^2-\frac{1}{3} \text{Tr}^2(\mathbf{T}_1-\mathbf{T}_2))}
\end{equation}
where $\|\mathbf{T}_1-\mathbf{T}_2\|_C =\sqrt{\text{Tr}(\mathbf{T}_1-\mathbf{T}_2)^2}$ is the Euclidean distance between two tensors $\{\mathbf{T}_1,\mathbf{T}_2\}$. $\text{Tr}\left(\cdot\right)$ means the trace of a matrix.

After the two normalized salient maps were achieved to indicate the local saliency structure distribution, JSM was builded by describing the matching degree between the two saliency maps at every pixel pairs in the overlapping regions of the two images. Given a point $\mathbf{x}_R$ in the reference image and its corresponding point $\mathbf{x}_M$ in the moving image after initial transformation, their joint-saliency value in a JSM is defined as
\begin{equation}
\begin{split}
\label{eq8}
& JS(\mathbf{x}_R,\mathbf{x}_M) \\
& =\min\{S_R(\mathbf{x}_R),S_M(\mathbf{x}_M)\}\frac{F\cdot G}{G+\|\text{LST}(\mathbf{x}_R)-\text{LST}(\mathbf{x}_M)\|_D} \\
\end{split}
\end{equation}
where $\{S_R(\cdot),S_M(\cdot)\}$ denote the saliency values in the salient map of a reference image and a moving one, $F=10$ and $G=\frac{1}{2}\max(\|\text{LST}(\mathbf{x}_R)-\text{LST}(\mathbf{x}_M)\|_D$ are two empirical parameters used to bound the final JSM values between 0 and 1. Note that it may introduce a situation that both of the corresponding pixels are assigned high saliency values in the structure-tensor based saliency maps, while their local variations of gradient orientations are in fact totally different. To avoid this situation, we also consider the dissimilarity measure between $\text{LST}(\mathbf{x}_R)$ and $\text{LST}(\mathbf{x}_M)$ at the denominator in equation (7).

Due to the JSM representing the degree of matching between the underlying saliency structure pairs, the mismatch scales should be inversely proportional to JSM values. Therefore, a zero or very small mismatch scale value is assigned to the corresponding structural regions with high JSM value in the multi-resolution registration context, while a large mismatch scale value is given to the unmatched structural regions with low value in JSM. Besides, owing to the low contribution to the nonrigid registration based on kernel regression, the mismatch scales in background or homogeneous regions are set to zero. According to the aforementioned idea, we defined the mismatch scale $\sigma_m$ as
\begin{eqnarray}
\sigma_m(\mathbf{x}) =
\begin{cases}
0,&if~ \mathbf{x}\in background~regions/\\&~~~~~~~homogeneous~regions
\cr 1/JS, &otherwise
\end{cases}
\end{eqnarray}
By this definition, the JSM map are transformed to the mismatch scale map that is used in the next step of JSS and local scale adaptive kernel regression.

\begin{figure}[!t]
\centering
\centerline{\includegraphics[width=3.5in]{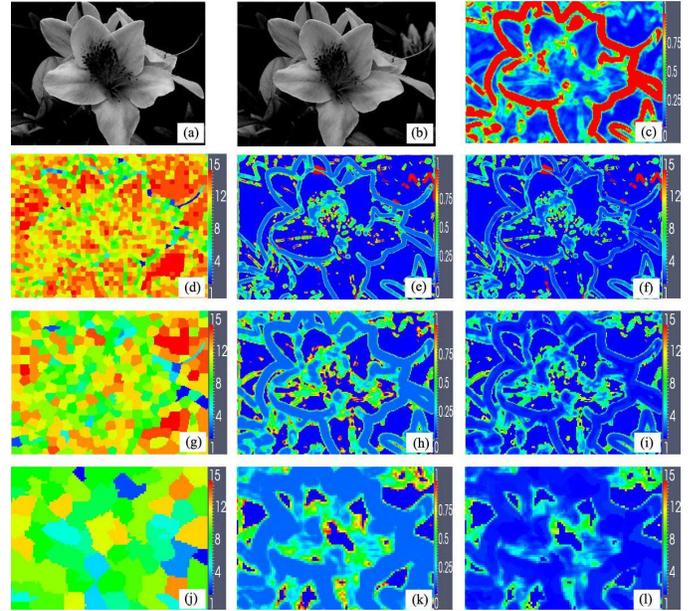}}
\caption{\small{\textbf{Flower images and their multi-resolution JSM, structure scale maps, mismatch scale maps and kernel scale maps.} (a)-(b) The reference and moving flower images at the $384\times 288$ pixels resolution. (c) JSM at the $192\times 144$ resolution, (d)-(f) structure scale, mismatch scale and kernel scale maps for the finest $384\times 288$ pixels resolution of images, (g)-(i) structure scale, mismatch scale and kernel scale maps for the $192\times 144$ pixels resolution of images. (j)-(l) structure scale, mismatch scale and kernel scale maps for the $96\times 72$ pixels resolution of images.}}
\label{fig2.eps}
\end{figure}

\subsection{Local Adaptive Scale for JSS Adaptive Kernel Regression}
Given the structure scale $\sigma_s$ and the mismatch scale $\sigma_m$, we are ready to design the local kernel scale $\sigma_d$ as
\begin{equation}
\begin{split}
\label{eq10}
\sigma_d = \max\{\sigma_s\times\sigma_m, 1\} \\
\end{split}
\end{equation}
where 1 avoids the local kernel scale being less than 1 pixel.

Fig. 2 illustrates the JSM, structure scale map, mismatch scale map and local kernel scale map at the multi-resolution scheme with the color scale representing different normalized joint saliency values or scale values. Fig. 2(a)-(b) are the reference and moving images with the $384\times 288$ pixels resolution. Fig. 2(c) is the JSM for the $192\times 144$ pixels resolution of images. Fig. 2(d)-(f) display the structure scale, mismatch scale and kernel scale maps for the finest resolution of images. Fig. 2(g)-(i) and Fig. 2(j)-(l) display the structure scale, mismatch scale and kernel scale maps for the $192\times 144$ and $96\times 72$ pixels resolutions of images, respectively.

From the left column of Fig. 2, the multi-resolution superpixel based structure scale map has demonstrated its success in determining the structure scales of the saliency structures in the multi-resolution images. Specifically, it roughly segments the foreground structural regions and does not segment the small structures at the coarse resolution (e.g., the stamen filament in the upper right corner of the reference image), so that there are more moderate structure scales (within the superpixels of large size) than the maximal and minimal scales being presented in the structure scale maps. With the increasing image resolution reducing the size of superpixels and enhancing the image details, the structure scale maps can precisely recognize the small structures (e.g., the small petals, the petal boundaries and the stamen filament at Fig. 2(g), (d)) such that the small structure scales are appropriately assigned to these small structures while the maximal structure scales being displayed at background or homogeneous regions.

The foreground large smooth structural regions in the coarse resolution initialize the sparse deformation vectors for subsequent multi-resolution nonrigid registration, while the small saliency structures displaying saliency details at the fine resolution refine the sparse deformation vectors. With these small structures being gradually joined in the iterative kernel regression, the background and homogeneous foreground regions achieve large structure scale, zero mismatch scale, and the smallest kernel scale (1) (Fig. 2(d)-(f) and Fig. 2(g)-(i)). The small saliency structures in the fine resolution have small structure scales, their final kernel scales are dependent on their mismatch scales (Fig. 2(e), (h), (k)); the locally matched small saliency structures with fine saliency details have large joint saliency values and very small mismatch scales, while the mismatched small saliency structures have very small joint saliency values and large mismatch scales (especially at the the missing correspondence and local large deformations in the upper right corner of images). Consequently, the matched small structures have small kernel scales while the mismatched saliency structures having relative large kernel scales in the fine resolution (see Fig. 2(f), (i), (l)). According to the aforementioned analysis in the multi-resolution scheme, the moving image's local saliency structures are gradually matched into the corresponding reference structures by iteratively selecting the structure scales and mismatch scales for the JSS \& local scale adaptive kernel regression.

\section{Experimental Results}
In this section, we use a set of 2D image pairs to validate the performance of the proposed method through comparing with our previous method \cite{12}, Advanced Normalized Tools (ANTs)\footnote{http://www.picsl.upenn.edu/ANTs} with elastic transformation and Mutual information (AMI) \cite{33}, AMI with cost-function Masking (AMM), Diffeomorphic Demons with Diffusion-like regularization (DDD) \cite{34} in Medical Image Processing, Analysis, and Visualization (MIPAV)\footnote{http://mipav.cit.nih.gov} and fast B-Spline with MI (BMI) in 3D Slicer\footnote{http://www.slicer.org}. The parameters of our two methods are: the number of pyramid levels is 5; the local similarity measure is mutual information. We set the parameters of AMI and AMM as: the histogram bin size is 32; the number of pyramid levels is 3; the iterations are set to $100\times100\times10$; the gradient step is 10; the default regularization is Gaussian filtering with a sigma of 3. The parameters of DDD method are set as follows: the variance of smoothing kernels is 2; the step scale is 1; the number of pyramid levels is 5; the number of iterations is 100. The parameters of the BMI method are selected as: the number of iterations is set to 100; the grid size is 15; the histogram bin size is 32; the spatial sample is 50000; the maximum deformation is 20. With those parameters all the methods mentioned above achieve their best performances.

\begin{figure}[!t]
\centering
\centerline{\includegraphics[width=3.5in]{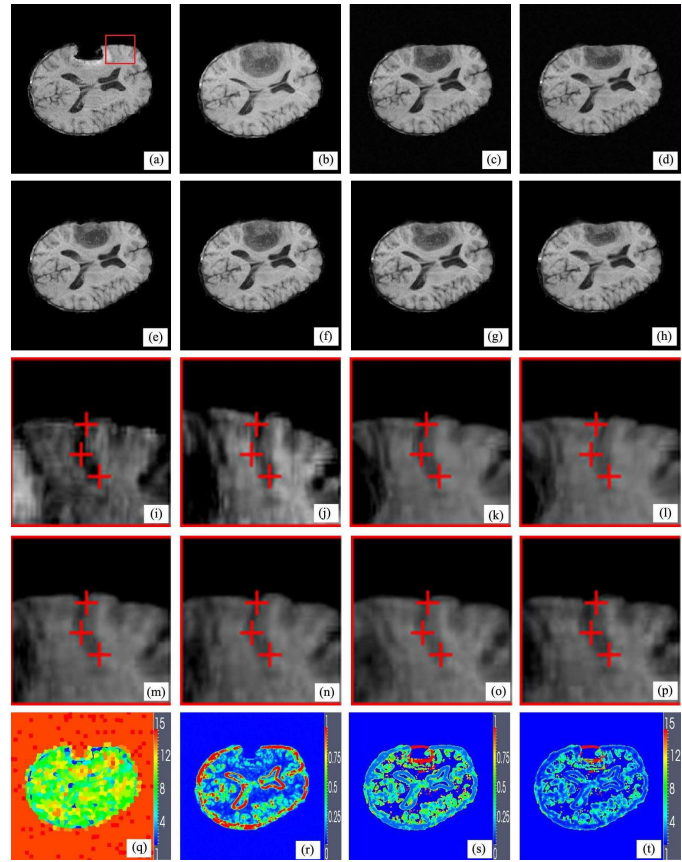}}
\caption{\small{\textbf{Brain tumor resection image registration.} (a)-(b) The reference and moving images, (c) the proposed method, (d) the previous method, (e) AMI, (f) AMM, (g) DDD, (h) BMI, (i)-(p) the same sulcus in (a)-(h) with desired spatial positions located by red cross, (q) structure scale map, (r) JSM, (s) mismatch scale map, (t) kernel scale map.}}
\label{fig3.eps}
\end{figure}

\begin{figure}[!t]
\centering
\centerline{\includegraphics[width=3.5in]{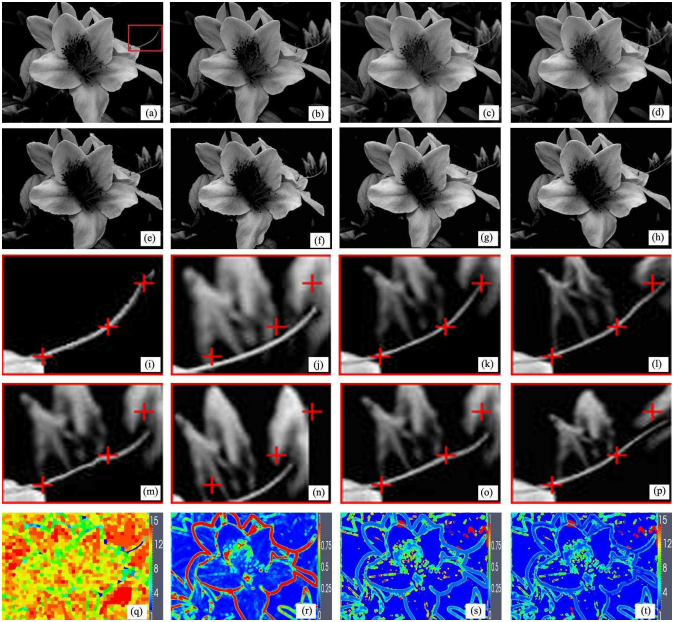}}
\caption{\small{\textbf{Flower image registration.} (a)-(b) The reference and moving images, (c) the proposed method, (d) the previous method, (e) AMI, (f) AMM, (g) DDD, (h) BMI, (i)-(p) the same stamen filament in (a)-(h) with desired spatial positions located by red cross, (q) structure scale map, (r) JSM, (s) mismatch scale map, (t) kernel scale map.}}
\label{fig4.eps}
\end{figure}

To evaluate the performance of the six competing methods, we not only zoom in some local small structures in the registered moving images and display their deviation from desired spatial positions located by several red crosses but also measure the registration errors at densely distributed landmarks selected by an expert. The landmark selection fully excludes outlier features while paying more attention to the identifiable locations at the local small saliency structures. Subsequently, the average landmark-based registration error distances and corresponding standard deviations of the six methods in every case are listed. Lower average error distances and lower standard deviation imply more accurate alignment of local structures.

The first experiment involves matching pre- and post-operative brain tumor resection images. Brain tissue severely suppressed by tumor in the preoperative image (Fig. 3(a)(b)) expands after tumor resection, which introduces not only the missing correspondence of tumor in the post-operative images but also the local large deformations caused by the brain shift. Fig. 3(c)-(h) are the registered results of the proposed method, the previous method, AMI, AMM, DDD and BMI. Visual inspection has revealed that the proposed method, the previous method, AMI and AMM methods apparently perform better than the DDD and BMI methods because the local brain deformation resulted from the latter two methods is either insufficient or somewhat excessive. The deformations of the sulcus near the missing corresponding tumor region in Fig. 3(a)-(h) are emphatically illustrated in Fig. 3(i)-(p). Comparing Fig. 3(k) with Fig.3(l) shows the improvement of the proposed method to the previous method in registration accuracy. The structure scale map, JSM, mismatch scale map and kernel scale maps at the finest resolution of the images for the proposed method are shown in Fig. 3(q)-(t).

\begin{figure}[!t]
\centering
\centerline{\includegraphics[width=3.5in]{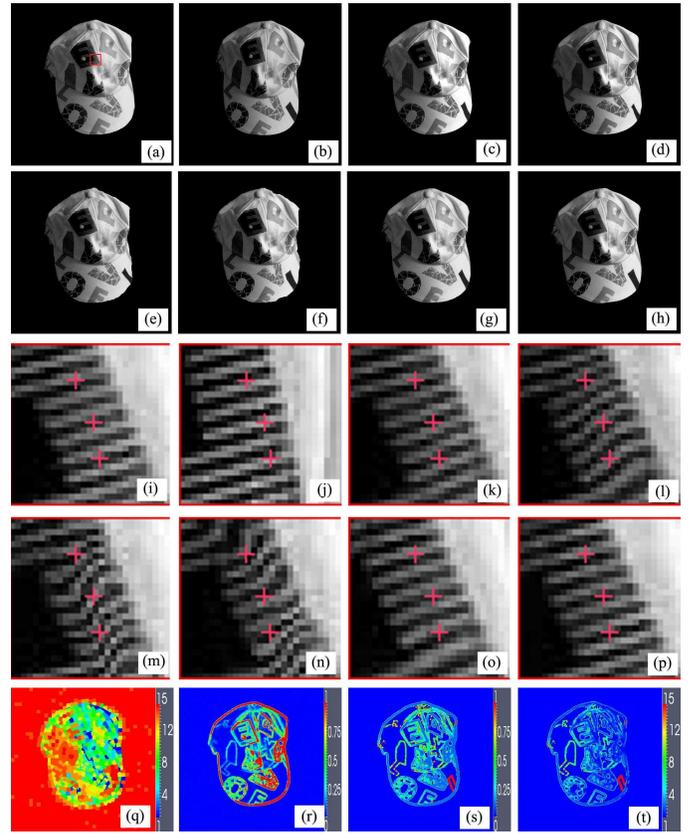}}
\caption{\small{\textbf{Hat image registration.} (a)-(b) The reference and moving images, (c) the proposed method, (d) the previous method, (e) AMI, (f) AMM, (g) DDD, (h) BMI, (i)-(p) the same stripes in (a)-(h) with desired spatial positions located by red cross, (q) structure scale map, (r) JSM, (s) mismatch scale map, (t) kernel scale map.}}
\label{fig5.eps}
\end{figure}

Fig. 4 displayed flower image registration cases, where the stamen filament in the right part of the reference image (Fig. 4(a)) is driven by the movement of the center flowers. In addition, some buds behind the stamen filament in the moving image (Fig. 4(b)) disappear in the reference image. A desired registered result of this case should properly deform the stamen filament according to the reference image regardless of the missing buds and flowers with large deformation. Fig. 4(c)-(h) show the registered results of the six methods, where the proposed method, our previous method and BMI outperform the other three method from the visual evaluation. However, the enlarged images (Fig. 4(i)-(p)) of the same stamen filament demonstrate that all the approaches except the proposed method fail to deform the stamen filament accurately. Fig. 4(p)-(t) show the structure scale map, JSM, mismatch scale map and kernel scale maps of the proposed method in this case.

A more challenging experiment is shown in Fig. 5. With the distortion of the hat, all the alphabets in the hat deform as well, especially the black stripes in 'E' having local large deformation. Besides, the missing 'I' in the reference image (Fig. 5(a)) appears in the moving image (Fig. 5(b)). The main difficulty in this experiment lies in the reasonable alignment of local small scale structures such as the stripes in 'E'. Because too many tiny structures are close to each other, mismatching in one structure will subsequently affect the deformation of its neighboring local tiny structures, and thus lead to poor structure alignment in a certain region. Fig. 5(c)-(h) show the registered moving images by the six methods. The enlarged images in the stripes of 'E' are displayed in Fig. 5(i)-(p). Comparatively, only the proposed and the BMI method preserve the matching accuracy of the stripes. The structure scale map, JSM, mismatch scale map and kernel scale maps of the proposed method are shown in Fig. 5(q)-(t).

Tab. 1 compares the average registration errors and the corresponding standard deviations of the 38 landmarks, 20 landmarks and 40 landmarks manually selected in the brain image, flower image and hat image registration. The proposed method has achieved sub-pixel registration accuracy for all the three experiments with the registration errors of (0.96$\pm$1.83, 0.93$\pm$3.01, 0.87$\pm$1.01), while the registration errors of the previous and AMI methods approximately keep (0.97$\pm$1.91, 1.14$\pm$2.96, 0.91$\pm$1.25) and (0.95$\pm$1.48, 1.69$\pm$3.49, 1.36$\pm$1.01). The other three methods can not perform well in these three challenging image registrations. In general, compared with the other state-of-the-art methods, only our method has achieved satisfying sub-pixel registration accuracy for all these challenging image registrations with outliers.

\begin{center}
\begin{table}
\centering
\renewcommand{\arraystretch}{0.6}
\caption{\small{\textbf{Landmark Registration errors (Mean+SD) of the six methods for the three grayscale image registrations} }}
\begin{tabular*}{0.5\textwidth}{@{\extracolsep{\fill}}llllllc}
\hline
proposed        &previous       &AMI                &AMM                &DDD                &BMI            \\
\hline
0.96$\pm$1.83   &0.97$\pm$1.91   &0.95$\pm$1.48     &1.16$\pm$1.99    &1.60$\pm$3.08   &1.15$\pm$1.89 \\
0.93$\pm$3.01   &1.14$\pm$2.96   &1.69$\pm$3.49     &8.38$\pm$8.82    &4.42$\pm$5.65   &1.55$\pm$3.49 \\
0.87$\pm$1.01   &0.91$\pm$1.25   &1.36$\pm$1.01     &2.75$\pm$3.63    &4.26$\pm$4.67    &2.50$\pm$4.05 \\
\hline
\end{tabular*}
\end{table}
\end{center}

\section{Conclusion}
In this paper, by designing the JSM-driven kernel scale estimator in aligning local structures with missing correspondence and local large deformation, we improve the previously proposed nonrigid registration using JSS adaptive kernel regression to achieve accurate local structure matching. Specifically, for every local structures to be registered, we combine their mismatch scale characterized by the JSM with their intrinsic structure scale so that the kernel function can adaptively control the size of sparse displacement vector samples participating in the JSS adaptive kernel regression for nonrigid image registration.

As indicated in \cite{5}, the iterative approach can improve the nonparametric estimates. In our work, the corresponding adaptive kernel scale selection, the kernel shape adaptivity and JSS-based weighted estimation \cite{13} iteratively deployed by the multi-resolution JSS adaptive kernel regression are significantly effective to improve the performance of local structure matching in the nonrigid image registration. The proposed method achieves the continuous and locally smooth transformation for accurately matching the local small structures with outliers and is favorably compared with state-of-the-art methods both in cases of missing correspondence and local large deformations. It is important to note that the computational cost of the proposed method is expensive even for 2D image registration, though it is easy to extend the proposed algorithm to 3D image registration. For reducing the computation burden of our approach even for the future 3D/4D nonrigid image registration, fast method is required to estimate the local adaptive structure scales, mismatch scale as well as discrete local structure-adaptive Gaussian kernels \cite{36} and implement structural adaptive kernel regression at every voxel.


%

\appendices


\section*{Acknowledgment}
The authors would like to thank Simon K. Warfield for providing MR brain images, E. Suarez-Santana for opening the source code in ITK project, R. Achanta and \emph{et al.} for opening the source code in SLIC. The authors thank the open source ANTs project, MIPAV, Diffeomorphid Demons, 3D Slicer and ITK project.

\ifCLASSOPTIONcaptionsoff
  \newpage
\fi


\begin{thebibliography}{}

\bibitem{1}
B. Qin, Z. Gu, X. Sun, Y. Lv, \emph{Registration of Images With Outliers Using Joint Saliency Map}, IEEE Signal Proc. Lett., 17(1): 91-94, 2010.

\bibitem{2}
A. S. El-Baz, R. Acharya U, M. Mirmehdi and \emph{et al.}, \emph{Multi Modality State-of-the-Art Medical Image Segmentation and Registration Methodoliges}, Springer, 2011.

\bibitem{3}
C. D. Lloyd, \emph{Local Models for Spatial Analysis}, Second Edition, CRC Press, Boca Raton, 2011.

\bibitem{4}
V. Katkovnik, A. Foi, K. Egiazarian, J. Astola, \emph{From Local Kernel to Nonlocal Multiple-Model Image Denoising}, Int. J. Comput. Vis., 86(1): 1-32, 2010.

\bibitem{5}
P. Milanfar, \emph{A Tour of Modern Image Filtering: New insights and methods, both practical and theoretical}, IEEE Signal Processing Magazine, 30(1): 106-128, 2013.

\bibitem{6}
D. Gerogiannis, C. Nikou, A. Likas, \emph{Registering sets of points using Bayesian regression}, Neurocomputing, 89, 122-133, 2012.

\bibitem{7}
A. Myronenko, X. Song, \emph{Point-set registration: coherent point drift}, IEEE Trans. Pattern Anal. Mach. Intell., 32, 2262-2275, 2010.

\bibitem{8}
B. Jian, B. C. Vemuri, \emph{Robust point set registration using Gaussian mixture models}, IEEE Trans. Pattern Anal. Mach. Intell., 33, 1633-1645, 2011.

\bibitem{9}
H. Chui, A. Rangarajan, \emph{A new point matching algorithm for non-rigid registration},  Comput. Vision Image Understanding, 89, 114-141, 2003.

\bibitem{10}
I. J. A. Simpson, J. A. Schnabel, A. R. Groves, J. L. R. Andersson, M. W. Woolrich, \emph{Probabilistic inferecence of regularisation in non-rigid registration}, NeuroImage, 59, 2438-2451, 2012.

\bibitem{11}
S.~Gao, L.~Zhang, H.~Wang, R.~de Crevoisier, D.~D.~Kuban, R.~Mohan, and L.~Dong, \emph{A deformable image registration method to handle distended rectums in prostate cancer radiotherapy}, Medical Physics, 33(9): 3304-3312, 2006.

\bibitem{12}
M. Brett, A. P. Leff, C. Rorden, and J. Ashburner, \emph{Spatial Normalization of Brain Images with Focal Lesion Using Cost Function Masking}, NeuroImage, 14(2): 486-500, 2001.

\bibitem{13}
http://arxiv.org/pdf/1302.0494.pdf

\bibitem{14}
E.~Suarez, C.-F.~Westin, E.~Rovaris, and J.~Ruiz-Alzola, \emph{Nonrigid registration using regularized matching weighted by local structure}, Medical Image Computing and Computer-Assisted Intervention-MICCAI 2002, LNCS 2489, pp.~581-589, 2002.

\bibitem{15}
H.~Knutsson and C.-F.~Westin, \emph{Normalized and differential convolution: Methods for interpolation and filtering of incomplete and uncertain data}, in Proc.~IEEE Computer Society Conf.~Computer Vision and Pattern Recognition, Jun.~16-19, pp.~515-523, 1993.

\bibitem{16}
E.~Ardizzone, R.~Gallea, O.~Gambino, and R.~Pirrone, \emph{Multi-modal Image Registration Using Fuzzy Kernel Regression}, Proceedings of the 16th IEEE international conference on Image processing, ICIP'09, pp.~193-196, 2009.

\bibitem{17}
B. Liu, J. Zhang, X. Liao, \emph{Projection Kernel Regression for Image Registration and Fusion in Video-Based Criminal Investigation}, Internation Conference on Multimedia and Signal Processing (CMSP'11), pp.~348-352, 2011.

\bibitem{18}
Jian Sun, Zongben Xu, \emph{Scale selection for anisotropic diffusion filter by Markov random field model}, Pattern Recognition, 43(8): 2630-2645, 2010.

\bibitem{19}
G. Gomez, J.L. Marroquin, L.E. Sucar, \emph{Probabilistic Estimation of Local Scale}, 15th International Conference on Pattern Recognition (ICPR'00), vol. 3, pp. 790-793, 2000.

\bibitem{20}
Lindeberg, \emph{Feature detection with automatic scale selection}, International Journal of Computer Vision, 30(2): 77-116, 1998.

\bibitem{21}
Comaniciu, \emph{An algorithm for data-driven bandwidth selection}, IEEE Trans. Pattern Anal. Mach. Intell., 25(2): 281-288, 2003.

\bibitem{22}
Adrian G. Bors, Nikolaos Nasios, \emph{Kernel Bandwidth Estimation for Nonparametric Modeling}, IEEE Transactions on Systems, Man, and Cybernetics, Part B: Cybernetics, 39(6): 1543-1555, 2009.

\bibitem{23}
M. Taron, N. Paragios, and M.-P. Jolly, \emph{Registration with Uncertainties and Statistical Modeling of Shape with Variable Metric Kernels}, IEEE Trans. Pattern. Anal. Mach. Intell., 31(1): 99-113, 2009.

\bibitem{24}
M. Hub, M. L. Kessler, and C. P. Karger, \emph{A Stochastic Approach to Estimate the Uncertainty Involved in B-Spline Image Registration}, IEEE Trans. Med. Imag., 28(11): 1708-1716, 2009.

\bibitem{25}
J. Kybic, \emph{Bootstrap Resampling for Image Registration Uncertainty Estimation Without Ground Truth}, IEEE Trans. Image Proc., 19(1): 64-73, 2010.

\bibitem{26}
P. Risholm, S. Pieper, E. Samset, and W. Wells, \emph{Summarizing and visualizing uncertainty in non-rigid registration}, in MICCAI 2010, ser. LNCS, T. Jiang, N. Navab, J. Pluim, and M. Viergever, Eds. Springer, Heidelberg, 6362, 554-561, 2010.

\bibitem{27}
T. Watanabe and C. Scott, \emph{Spatial Confidence Regions for Quantifying and Visualizing Registration Uncertainty}, B.M. Dawant et al. (Eds.): WBIR 2012, LNCS 7359, pp. 120-130, 2012.

\bibitem{28}
I. J. A. Simpson, M. W. Woolrich, J. L. R. Andersson, A. R. Groves, J. A. Schnabel and the Alzheimer's Disease Neuroimaging Initiative, \emph{Ensemble Learing Incorporating Uncertain Registration}, IEEE Trans. Med. Imag., preprint 2013. DOI: 10.1109/TMI.2012.2236651

\bibitem{29}
A. Borji, L. Itti, \emph{State-of-the-Art in Visual Attention Modeling}, IEEE Trans. Pattern. Anal. Mach. Intell., 35(1): 185-207, 2013.

\bibitem{30}
R. Achanta, A. Shaji, K. Smith, A. Lucchi, P. Fua, and S. Susstrunk, \emph{SLIC Superpixels Compared to State-of-the-art Superpixel Methods}, IEEE Trans. Pattern Anal. Mach. Intell., 34(11): 2274-2281, 2012.

\bibitem{31}
P. Perona, J. Malik, \emph{Scale-space and edge detection using anisotropic diffusion}, IEEE Trans. Pattern Anal. Mach. Intell., 12(7): 629-639, 1990.

\bibitem{32}
H.~Zhang, P. A. Yushkevich, D. C. Alexander, J. C. Gee, \emph{Deformable registration of diffusion tensor MR images with explicit orientation optimization}, Med. Image Anal., 10(5): 764-785, 2006.

\bibitem{33}
B. B. Avants, N. J. Tustison, G. Son, P. A. Cook, A. Klein, J. C. Gee, \emph{A reproducible evaluation of ANTs similarity metric performance in brain image registration}, NeuroImage, 54(3): 2033-2044, 2011.

\bibitem{34}
T. Vercauteren, X. Pennec, A. Perchant, N. Ayache, \emph{Diffeomorphic demons: Efficient non-parametric image registration}, NeuroImage, 45, S61-S72, 2009.

\bibitem{35}
J. Rissanen, \emph{A universal prior for integers and estimation by Minimum Description Length}, Annals of Statistics, 11(2): 416-431, 1983.

\bibitem{36}
K. N. Chaudhury, S. Sanyal, \emph{Improvements on "Fast Space-Variant Elliptical Filtering Using Box Splines"}, IEEE Trans. Image Proc., 21(9): 3915-3923, 2012.












\end{thebibliography}
\end{document}